%% file: main.tex
\documentclass[10pt,twocolumn,letterpaper]{article}

\usepackage{iccv}              
\usepackage{color}
\usepackage{multicol}
\usepackage{multirow}
\usepackage{booktabs}
\usepackage{threeparttable}
\usepackage{array}
\usepackage{booktabs}
\usepackage{bbding}
\usepackage{diagbox}
 \usepackage{graphicx}
 \usepackage{amsmath}
  \usepackage{amsfonts}
  \usepackage[mathscr]{euscript}
  \usepackage{pifont}

\definecolor{iccvblue}{rgb}{0.21,0.49,0.74}
\usepackage[pagebackref,breaklinks,colorlinks,allcolors=iccvblue]{hyperref}

\def\paperID{11213} 
\def\confName{ICCV}
\def\confYear{2025}

\title{OS-DiffVSR: Towards One-step Latent Diffusion Model for High-detailed Real-world Video Super-Resolution}

 \author{Hanting~Li$^{\ast}$, Huaao~Tang\thanks{Equal contribution}, Jianhong~Han, Tianxiong~Zhou, Jiulong~Cui, Haizhen~Xie, Yan~Chen, Jie~Hu\thanks{Corresponding author}\\
Huawei Noah’s Ark Lab}

\begin{document}

\maketitle
\input{sec/0_abstract}    
\input{sec/1_intro}
\input{sec/2_formatting}

\input{sec/3_method}

\input{sec/4_experiments}

\input{sec/5_conclusion}

{
    \small
    \bibliographystyle{ieeenat_fullname}
    \bibliography{main}
}

\end{document}

%% file: sec/0_abstract.tex
\begin{abstract}
Recently, latent diffusion models has demonstrated promising performance in real-world video super-resolution (VSR) task, which can reconstruct high-quality videos from distorted low-resolution input through multiple diffusion steps. 
Compared to image super-resolution (ISR), VSR methods needs to process each frame in a video, which poses challenges to its inference efficiency. However, video quality and inference efficiency have always been a trade-off for the diffusion-based VSR methods. In this work, we propose \textbf{O}ne-\textbf{S}tep \textbf{Diff}usion model for real-world \textbf{V}ideo \textbf{S}uper-\textbf{R}esolution, namely OS-DiffVSR. Specifically, we devise a novel adjacent frame adversarial training paradigm, which can significantly improve the quality of synthetic videos. Besides, we devise a multi-frame fusion mechanism to maintain inter-frame temporal consistency and reduce the flicker in video. Extensive experiments on several popular VSR benchmarks demonstrate that OS-DiffVSR can even achieve better quality than existing diffusion-based VSR methods that require dozens of sampling steps.
\end{abstract}

%% file: sec/1_intro.tex
\section{Introduction}
\label{sec:intro}

With rapid iteration of video capture devices and improvement of Internet transmission speed, video content plays an increasingly important role in the digital media industry. The video super-resolution (VSR) task aims to convert low-resolution (LR) videos into high-resolution (HR) videos with rich details, thereby greatly improving video quality. 
real-world video degradations are very complex, including noise, blur, compression artifact, and other more complex compound degradations, which pose great challenges to VSR task.

\begin{figure}[t]

  \centering
  \includegraphics[width=0.9\linewidth]{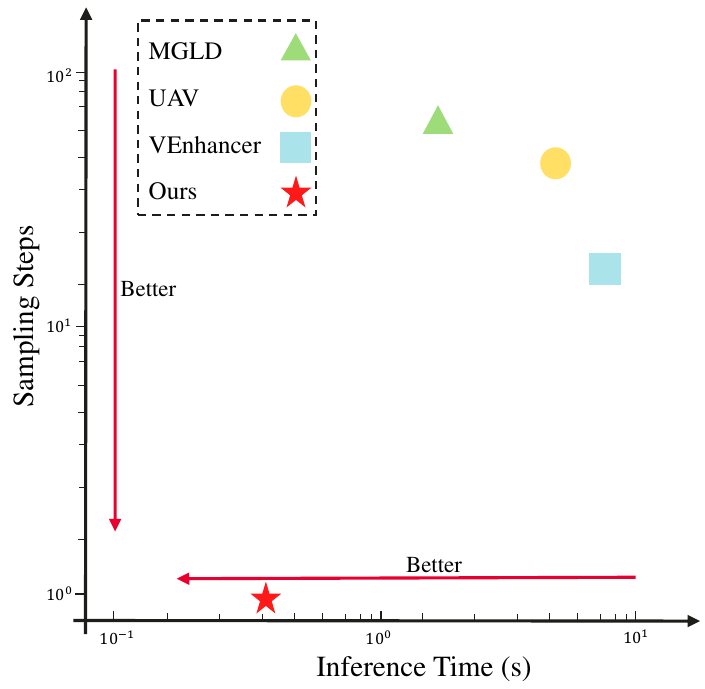}
  \caption{Comparison of model efficiency between existing diffusion-based VSR methods and our OS-DiffVSR.}\vspace{-0.5cm}
 \label{fig:1}
\end{figure}

Compared to image super-resolution (ISR), VSR methods needs to process each frame in a video sequence, which puts higher requirements on the model efficiency. VSR methods based on convolutional neural networks (CNNs) \cite{chan2021basicvsr, wang2021real,tian2020tdan} and recursive models \cite{liang2022recurrent,fuoli2019efficient,haris2019recurrent} usually have higher reasoning efficiency, but are difficult to generate videos with rich details. Recently, diffusion models \cite{ho2020denoising,rombach2022high,blattmann2023align} have achieved remarkable success in various image and video synthesis tasks, such as text to video synthesis \cite{esser2023structure,chen2023motion}, video editing \cite{ceylan2023pix2video,chai2023stablevideo}, and video restoration \cite{li2025tdm,zhou2024upscale,yang2024motion}. These methods can synthesize realistic videos through dozens or even hundreds of diffusion sampling steps, which makes the generation process time-consuming and difficult to meet the real-time requirements in many practical application scenarios. 

Recently, compressing the sampling steps of vanilla diffusion models \cite{ho2020denoising} based on stochastic differential equations (SDE) to fewer steps or even one step through distillation \cite{yin2024one} or ordinary differential equation-based methods \cite{liu2022flow} has become the mainstream technical route to reduce the inference time. However, the performance of one-step diffusion-based methods still lag behind that of the multi-step methods.

Therefore, we propose \textbf{O}ne-\textbf{S}tep \textbf{Diff}usion model for real-world \textbf{V}ideo \textbf{S}uper-\textbf{R}esolution (OS-DiffVSR) to extend the one-step ISR method (i.e., OSEDiff \cite{wu2025one}) to VSR task. OS-DiffVSR greatly improves the quality of synthetic videos under one-step sampling setting, bringing the diffusion-based VSR method one step closer to practical applications. As shown in Figure~\ref{fig:1}, the proposed OS-DiffVSR has obvious advantages in both dimensions of inference time and sampling step when comparing with other multi-step diffusion-based VSR methods. Specifically, we devise a novel adjacent frame adversarial training (AFAT) paradigm, which trains the discriminator by conducting contrastive learning between adjacent frames. Under this training paradigm, OS-DiffVSR can even exceeding the performance of multi-step diffusion-based VSR methods. Besides, the proposed multi-frame fusion (MFF) module can improve the temporal consistency of the synthetic videos. In summary, our contributions are as follows:

\begin{itemize}
    \item We proposed OS-DiffVSR, a one-step diffusion-based VSR method. OS-DiffVSR can produce compelling results comparable to multi-step diffusion models with one-step sampling.
    \item A novel adjacent frame adversarial training paradigm is proposed to improve the visual quality under one-step sampling setting. 
    \item We devise a multi-frame fusion module to improve the temporal consistency of the synthetic videos.
    \item Extensive experiments on five popular VSR datasets demonstrate that OS-DiffVSR exceeds existing VSR methods. The effectiveness of our method is further corroborated by the qualitative comparisons. 
\end{itemize}

%% file: sec/2_formatting.tex
\section{Related Work}
\label{sec:relatedwork}

\subsection{Video Super-Resolution} 

Traditional video super-resolution (VSR) methods have evolved from convolutional neural networks (CNNs) to transformers for handling temporal information. Early approaches predominantly relied on optical flow estimation\cite{kim2018spatio,sajjadi2018frame,xue2019video,chan2021basicvsr} or deformable convolutions \cite{tian2020tdan,wang2019edvr,jo2018deep,chan2021understanding} for feature alignment and fusion. Later developments incorporated attention mechanisms \cite{liang2024vrt, liang2022recurrent,cao2021video} to facilitate information communication and integration within the network. However, these methods typically trained under simplified degradation models, resulting in limited generalization capabilities and a significant performance drop when confronted with complex real-world degradations. Subsequent research has focused on operating real-world VSR, as exemplified by DBVSR\cite{pan2021deep}, RealBasicVSR\cite{chan2022investigating}, FastRealVSR\cite{xie2023mitigating}, and RealViFormer\cite{zhang2024realviformer}.

Nevertheless, these approaches lack generative priors, which hinders their ability to synthesize sharp and realistic textures, ultimately constraining their overall performance. Recent progress in VSR has seen the rise of diffusion prior-based methods, which have effectively addressed the limitations of traditional approaches in terms of detail sharpness and realism. However, a considerable disadvantage of almost all these methods is their dependence on iterative sampling procedures, leading to prohibitively high computational costs during inference. Our proposed method, conversely, harnesses the generative capabilities of diffusion models while necessitating only a single sampling step, achieving optimal inference efficiency without compromising generation quality.

\subsection{Diffusion Prior for Video Super-Resolution}

Employing diffusion-based generative priors in video super-resolution tasks allows for the synthesis of more authentic and refined details, ultimately contributing to an improved visual experience. A number of recent approaches have investigated the application of diffusion priors to video super-resolution. These approaches can be broadly categorized into two groups: those leveraging Text-to-Image (T2I) priors and those leveraging Text-to-Video (T2V) priors.


Many studies leverage pre-trained T2I models (e.g., Stable Diffusion\cite{rombach2021highresolution} and its variants) as backbones and build upon them to adapt to the video super-resolution task. Common techniques employed include:
\begin{itemize}
\item Utilizing image super-resolution models as backbone: StableVSR \cite{rota2024enhancing}, Upscale-A-Video \cite{zhou2024upscale}, DiffVSR \cite{li2025diffvsr} and others are based on the SD$\times$4 Upscaler\cite{Rombach_2022_CVPR}.
\item Incorporating motion information guidance \cite{rota2024enhancing,zhou2024upscale}: Extracting motion information via optical flow and integrating it into the diffusion sampling process to ensure temporal consistency. 
\item Constructing temporal modules\cite{li2025diffvsr,chen2024learning,yang2024motion}: Adding 3D convolutions, temporal attention mechanisms, etc., to pre-trained T2I models, such as U-Nets and VAE decoders, to explicitly model temporal correlations in videos. 
\item Adopting advanced architectures: SeedVR \cite{wang2025seedvr}, based on SD3\cite{esser2024scaling}, employs shifted window based MM-DIT and causal video VAE, achieving effective super-resolution for arbitrary-length and resolution videos.
\item Training-Free Methods: DiffIR2VR-zero \cite{yeh2024diffir2vr} proposes a training-free method that achieves VSR through optical flow-guided hierarchical latent warping and hybrid flow-guided spatial-aware token merging.
\end{itemize}
In summary, VSR methods based on T2I priors capitalize on the powerful generative capabilities, supplemented by various optimization techniques tailored for video data, thereby achieving a good performance. A limitation of these approaches is their lack of optimization for inference efficiency, as they generally necessitate 30 to 50 sampling steps. 
The method proposed in this paper also falls into this category. But we utilize the research results on step distillation in the T2I community, achieving single-step video super-resolution, result in a substantial improvement in inference efficiency.


Another approach is to directly leverage T2V models as priors. VEnhancer \cite{he2024venhancer} and STAR \cite{xie2025star} are both based on I2VGen-XL\cite{zhang2023i2vgen} and build upon it by adding 3D ControlNet\cite{zhang2023adding}. VEnhancer\cite{he2024venhancer} focuses primarily on enhancing the resolution of AIGC videos. STAR\cite{xie2025star} utilizes local information enhancement modules and dynamic frequency loss to improve real-world video super-resolution performance.
However, due to the immense parameter size of T2V models, methods based on T2V priors face challenges in terms of inference speed, with their inference latency typically higher than that of methods based on T2I models. Therefore, reducing the inference cost of T2V models is one of the future research directions in this field.

%% file: sec/3_method.tex
\section{Method}

\begin{figure*}[t]

  \centering
  \includegraphics[width=\linewidth]{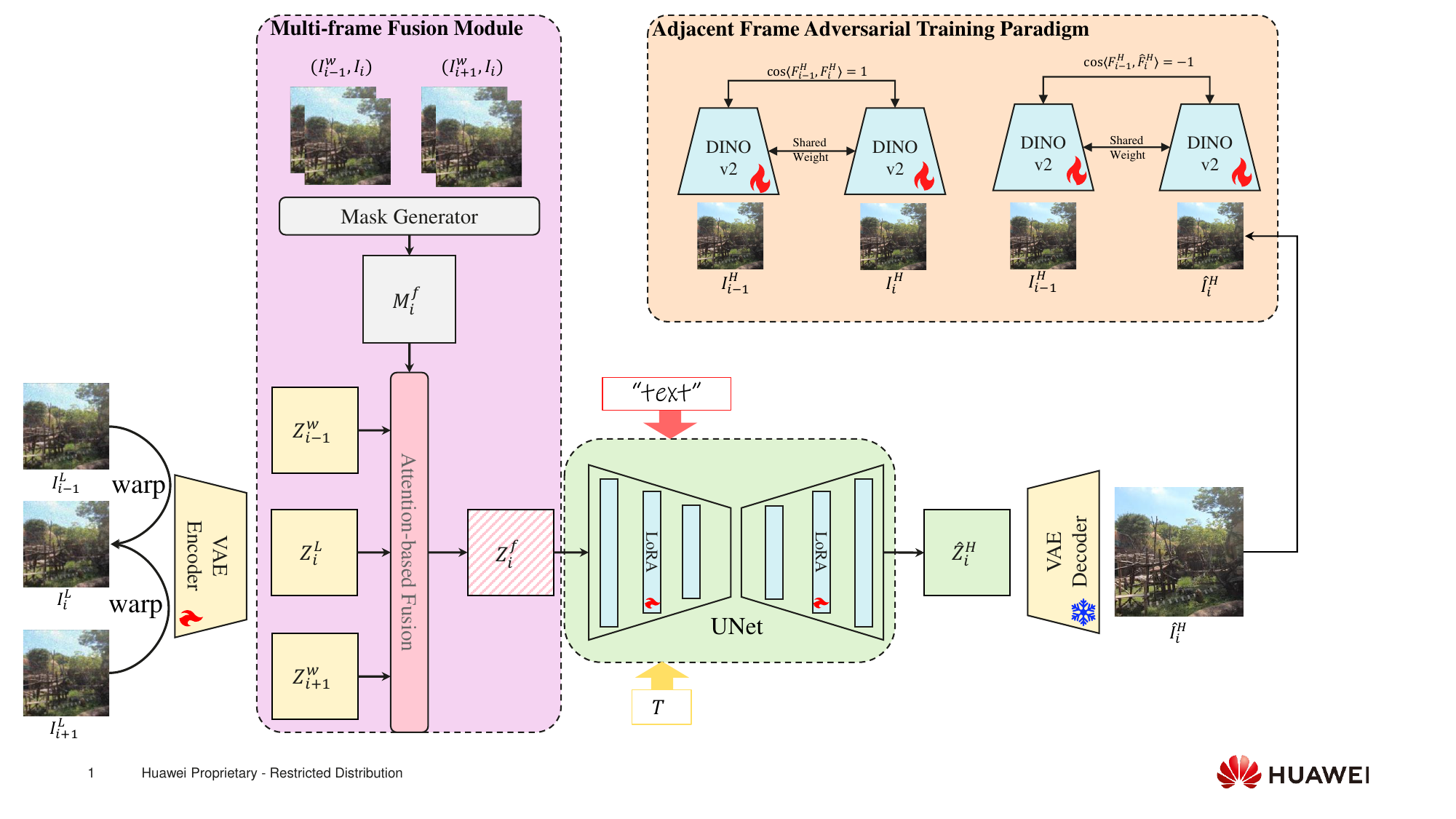}
  \caption{Overview of the proposed OS-DiffVSR. The overall framework mainly contains a generator composed of multi-frame fusion module and UNet and a adjacent frame discriminator based on DINOv2 \cite{oquab2023dinov2}.}
 \label{fig:pipline}
\end{figure*}

Extending powerful and stable image synthesis diffusion
model by adding trainable parameters has become a common paradigm for building VSR methods \cite{zhou2024upscale,yang2024motion}. Inspired by this idea, we built our method based on the latest ISR method (i.e., OSEDiff \cite{wu2025one}). In this work we mainly focus on tackling two main challenges: 1) improving the \textit{video quality} under one-step sampling conditions, 2) improving the \textit{temporal consistency} of synthesized videos.

To address the above two challenges, we proposed OS-DiffVSR. As shown in Figure~\ref{fig:pipline}, OS-DiffVSR mainly consists of three key components, which are multi-frame fusion module, Denoising UNet, and the adjacent frame adversarial training paradigm. For a given low-resolution video sequence consisting of $N$ frames $\mathscr{V}_{LR} = \{I^L_1, I^L_2, \cdot\cdot\cdot, I^L_N\}$, Our goal is to reconstruct the corresponding high-resolution video $\mathscr{\hat{V}}_{HR} = \{\hat{I}^H_1, \hat{I}^H_2, \cdot\cdot\cdot, \hat{I}^H_N\}$ from $\mathscr{V}_{LR}$. Firstly, we use the VAE encoder to map each frame into the latent space. For each latent feature $Z^L_i$, We fuse the information of warped adjacent frames (i.e., $Z^w_{i-1}$ and $Z^w_{i+1}$) through the proposed MFF module. Since the frame fusion is completed before being fed to UNet, the redundant computational overhead of intermediate-layer fusion is avoided, which allows OS-DiffVSR to fuse temporal information with minimal computational overhead. Then the fused latent embed $Z^f_{i}$ is fed into the denoising UNet together with the text prompt and the fixed time step $T$, which is consistent with the noise prediction process in OSEDiff. The denoised latent feature $\hat{Z}_i^H$ is decoded by the frozen VAE decoder to obtain the high-resolution output $\hat{I}_i^H$. In addition to the mean square error (MSE) loss and perceptual loss commonly used in super-resolution tasks, we also propose a novel adjacent frame adversarial training paradigm to guide the generator to synthesize more realistic details through the previous frame.

During the training process, the VAE encoder and UNet are fine-tuned with low-rank adaptation (LoRA), while the multi-frame fusion module and the discriminator (i.e., DINOv2 \cite{oquab2023dinov2}) are trained under full-parameter tuning setting. Only the VAE decoder is frozen throughout the training process.

\subsection{Preliminary: Latent Diffusion Model}
Diffusion models \cite{ho2020denoising} are a class of generative models that convert Gaussian noise into specific data samples through an iterative inverse Markov process. While the latent diffusion models (LDMs) \cite{rombach2022high} first map the data into the low-dimensional latent space, and then completes the diffusion and denoising process in the latent space. This paradigm has greatly promoted the development of high-dimensional data (e.g., images and video) generation \cite{blattmann2023align,podell2023sdxl}.

During the training stage, given a latent feature $Z^L$$\sim$$p_{LR}$ of a low-resolution image. The noised latent feature $Z^L_{t}$ at the $t$-th diffusion step can be calculated by,
\begin{equation}\label{eq:1}
Z^L_{t}=\alpha_tZ^L+\beta_t\epsilon,
\end{equation}
\noindent where $\alpha_t$ and $\beta_t$, $t$$\in$$\{1,2,\cdot\cdot\cdot, T\}$ define a noise schedule. Follow the one-step sampling process in \cite{wu2025one}, the denoised latent feature $\hat{Z}^H$ can be obtained by a UNet $\epsilon_\theta$ that can predict the noise in $Z^L_{t}$, which can be formulated as,
\begin{equation}\label{eq:2}
\hat{Z}^H = \frac{Z^L-\beta_T\epsilon_\theta(Z^L;T,c)}{\alpha_T}.
\end{equation}
\noindent Where $c$ denotes text descriptions corresponding to each frame, and $T$ represents the total number of diffusion steps.

\subsection{Adjacent Frame Adversarial Training Paradigm}
Adversarial training is a common strategy to improve performance of super-resolution models \cite{wang2020deep}. By training a discriminator to distinguish from the ground truth and generated HR results, the discriminator can help the SR model synthesize more high-detailed images \cite{wang2021real}. Although vanilla adversarial training paradigm has achieved promising result in the super-resolution tasks, in the VSR task, it can still be improved in the following aspects: 1) vanilla adversarial training paradigm do not fully exploit the information from other frames especially adjacent frame; 2) Traditional discriminator can only implicitly optimize the generator by narrowing the gap between real data and synthetic data distribution, but cannot provide direct guidance for specific samples. Therefore, we propose a novel adjacent frame adversarial training paradigm to train our OS-DiffVSR. We choose DINOv2 \cite{oquab2023dinov2}, a foundation model widely used in computer vision tasks, as our discriminator.

\paragraph{Discriminator:} The optimization goal of our discriminator is to distinguish the super-resolution result of the current frame $\hat{I}^H_i$  from ground truth $I^H_i$ based on the previous frame. By conducting contrastive learning between real adjacent frame pairs (i.e., $I_{i-1}^H$ and $I_{i}^H$) and fake adjacent frame pairs (i.e., $I_{i-1}^H$ and $\hat{I}_{i}^H$), the discriminator use the previous frame as a reference to distinguish the synthesized video, thus guiding the generator to generate a more realistic video. Specially, by utilizing the every output token of DINOv2, we can perform adversarial training at patch level, and the loss function of the discriminator can be defined as,

\begin{equation}\label{eq:3}\footnotesize
\mathscr{L}_{D}=-\frac{1}{PQ}\sum_{p=1}^{P}\sum_{q=1}^{Q}\log\frac{e^{\langle F_{i-1,p,q}^H,  F_{i,p,q}^H \rangle/\tau}}{e^{\langle F_{i-1,p,q}^H,  F_{i,p,q}^H \rangle/\tau}+e^{\langle F_{i-1,p,q}^H,  \hat{F}_{i,p,q}^H \rangle/\tau}},
\end{equation}
\noindent with
\begin{equation}\label{eq:4}
\begin{aligned}
&F_{i}^H = \mathscr{F}_{\theta}(I_i^H),\\
&\hat{F}_{i}^H = \mathscr{F}_{\theta}(\hat{I}_i^H).
\end{aligned}
\end{equation}

\noindent Where $\langle\cdot, \cdot\rangle$ stands for the cosine similarity between two vectors and $\tau$ denotes the temperature parameter used to control the margin between real and fake frame pairs. $I_i^H$ and $\hat{I}_i^H$ represent the $i$-th frame of the ground truth video and the synthetic video, respectively. $\mathscr{F}_{\theta}$ is our discriminator DINOv2 and $F_{i,p,q}^H$ is the patch-level feature at saptial position $(p,q)$ extracted by the discriminator.

\paragraph{Generator:} The training objective of the generator is to maximize the similarity between the synthetic frame and the real adjacent frame. Therefore, the loss function of the generator can be written as,
\begin{equation}\label{eq:5}
\mathscr{L}_{G}=-\frac{1}{PQ}\sum_{p=1}^{P}\sum_{q=1}^{Q}\langle F_{i-1,p,q}^H,  \hat{F}_{i,p,q}^H \rangle.
\end{equation}

\begin{figure}[t]

  \centering
  \includegraphics[width=\linewidth]{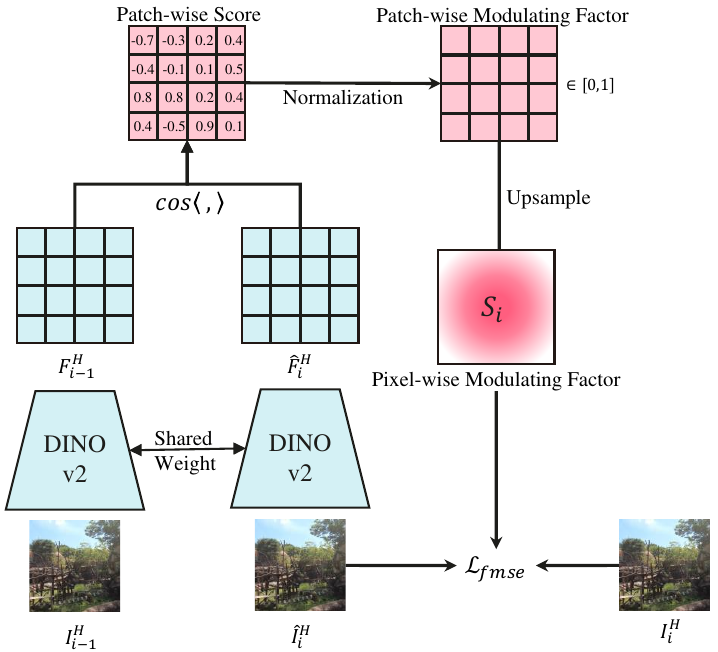}
  \caption{Illustration of the proposed focal mean square error.}
 \label{fig:focal}
\end{figure}

\paragraph{Focal Mean Square Error}
Since our discriminator can perform patch-level scoring based on the cosine similarity between $I_i^H$ and $\hat{I}_i^H$, we further use these scores to locate spatial locations with poor visual quality, thereby enhancing the reconstruction constraints of these spatial regions. Inspired by the focal loss \cite{lin2017focal} that can make the network pay more attention to the classes with less samples, we propose focal mean square loss to make the model focus on optimizing areas where the visual quality is poor. Based on the above insights, the focal mean square (FMSE) loss can be defined as,

\begin{equation}\label{eq:6}
\mathscr{L}_{fmse}=S_i^{\gamma}\left \| I_{i}^H-\hat{I}_{i}^H \right \|_2,
\end{equation}
\noindent with
\begin{equation}\label{eq:7}
S_i = \frac{1-\langle F_{i-1}^H,  \hat{F}_{i}^H \rangle}{2}\uparrow_{\times n}.
\end{equation}
\noindent Where $S_i$$\in$$[0,1]$ denotes the pixel-wise modulating factor of the $i$-th frame and $\gamma$$\ge$$0$ represent the focusing parameter. $\uparrow_{\times n}$ is $n$ times upsampling, which is used to upscale the patch-level feature to the same size as the original image resolution ($n$$=$$14$ in our work).

\subsection{Multi-frame Fusion Module}
The most significant difference between ISR and VSR is that VSR methods can obtain the temporal information from adjacent frames (e.g., previous frame and successor frame). Making full use of these information can not only improve the temporal consistency, but also improve the visual quality. In order to better fuse multi-frame information, we first utilize optical flow to align adjacent frames to the current frame:
\begin{equation}\label{eq:8}
\begin{aligned}
&I^w_{i-1} =\rm Warp\it(I^L_{i-1},OF^L_{b,i-1}),\\
&I^w_{i+1} =\rm Warp\it(I^L_{i+1},OF^L_{f,i+1}).
\end{aligned}
\end{equation}
\noindent Where $OF^L_{b,i}$ and $OF^L_{f,i}$ are the backward and forward optical flow of the $i$-th frame computed on the low-resolution sequence, respectively. Then the VAE encoder $E_{vae}$ is used to map the warped frame together with the current frame into the latent space:
\begin{equation}\label{eq:9}
\begin{aligned}
&Z^w_{i-1} = E_{vae}(I^w_{i-1}),\\
&Z^w_{i+1} = E_{vae}(I^w_{i+1}),\\
&Z^L_{i} = E_{vae}(I^L_{i}).
\end{aligned}
\end{equation}

\begin{table*}[t!]\footnotesize
\begin{center}
\begin{tabular}{c|c|cccccccc}
\toprule
Datasets&Metrics&Real-ESRGAN\cite{wang2021real}&SD $\times$4 Upscaler\cite{Rombach_2022_CVPR}&RealBasicVSR\cite{chan2022investigating}&MGLD\cite{yang2024motion}&VEnhancer\cite{he2024venhancer}&UAV\cite{zhou2024upscale}&Ours\cr
\midrule
\multirow{5}{*}{REDS4}&PSNR$\uparrow$&22.225&21.511&\textcolor{red}{23.479}&\textcolor{blue}{22.780}&18.907&21.889&21.789\cr
                      &CLIP-IQA$\uparrow$&\textcolor{blue}{0.465}&0.193&0.379&0.411&0.218&0.251&\textcolor{red}{0.544}\cr
                       &MUSIQ$\uparrow$&\textcolor{blue}{66.601}&27.901&65.259&66.476&41.9984&52.372&\textcolor{red}{67.286}\cr
                       &MANIQA$\uparrow$&0.609&0.357&0.623&\textcolor{blue}{0.645}&0.4799&0.487&\textcolor{red}{0.646}\cr
                        &DOVER$\uparrow$&0.617&0.264&\textcolor{blue}{0.631}&\textcolor{red}{0.697}&0.603&0.481&0.627\cr
\midrule
\multirow{5}{*}{UDM10}&PSNR$\uparrow$&26.428&24.535&\textcolor{red}{27.756}&\textcolor{blue}{26.594}&20.027&25.768&25.308\cr
                      &CLIP-IQA$\uparrow$&\textcolor{blue}{0.502}&0.240&0.469&0.500&0.376&0.378&\textcolor{red}{0.619}\cr
                       &MUSIQ$\uparrow$&62.429&32.325&62.835&\textcolor{blue}{62.923}&55.653&54.275&\textcolor{red}{66.071}\cr
                       &MANIQA$\uparrow$&0.552&0.350&0.564&\textcolor{blue}{0.569}&0.497&0.460&\textcolor{red}{0.579}\cr
                        &DOVER$\uparrow$&0.776&0.330&\textcolor{blue}{0.784}&0.769&0.714&0.667&\textcolor{red}{0.785}\cr
\midrule
\multirow{5}{*}{SPMCS}&PSNR$\uparrow$&22.059&21.337&\textcolor{red}{22.961}&\textcolor{blue}{22.357}&17.196&21.366&21.347\cr
                      &CLIP-IQA$\uparrow$&\textcolor{blue}{0.492}&0.238&0.442&0.469&0.343&0.419&\textcolor{red}{0.585}\cr
                       &MUSIQ$\uparrow$&66.910&33.183&\textcolor{red}{67.576}&\textcolor{blue}{67.401}&51.986&61.052&66.322\cr
                       &MANIQA$\uparrow$&0.574&0.381&0.574&\textcolor{blue}{0.610}&0.533&0.510&\textcolor{red}{0.623}\cr
                        &DOVER$\uparrow$&\textcolor{red}{0.675}&0.303&0.604&0.628&0.449&0.600&\textcolor{blue}{0.664}\cr
\midrule
\multirow{5}{*}{YouHQ40}&PSNR$\uparrow$&\textcolor{blue}{23.485}&21.448&\textcolor{red}{23.910}&23.332&18.824&23.035&22.719\cr
                      &CLIP-IQA$\uparrow$&0.468&0.289&0.482&\textcolor{blue}{0.543}&0.380&0.434&\textcolor{red}{0.598}\cr
                       &MUSIQ$\uparrow$&59.719&28.392&64.336&\textcolor{red}{65.626}&49.137&54.751&\textcolor{blue}{65.142}\cr
                       &MANIQA$\uparrow$&0.507&0.328&0.546&\textcolor{blue}{0.579}&0.442&0.472&\textcolor{red}{0.583}\cr
                        &DOVER$\uparrow$&\textcolor{red}{0.860}&0.509&0.836&\textcolor{blue}{0.858}&0.754&0.767&0.829\cr
\midrule
\midrule
\multirow{5}{*}{VideoLQ}&NIQE$\downarrow$&4.081&8.560&3.702&\textcolor{blue}{3.652}&5.460&5.426&\textcolor{red}{3.455}\cr
                      &CLIP-IQA$\uparrow$&0.377&0.267&\textcolor{blue}{0.381}&0.366&0.288&0.212&\textcolor{red}{0.575}\cr
                       &MUSIQ$\uparrow$&56.240&23.748&\textcolor{blue}{59.805}&56.528&43.751&38.711&\textcolor{red}{65.071}\cr
                       &MANIQA$\uparrow$&0.507&0.321&\textcolor{blue}{0.548}&0.536&0.415&0.379&\textcolor{red}{0.585}\cr
                        &DOVER$\uparrow$&0.744&0.345&\textcolor{blue}{0.749}&0.747&0.655&0.556&\textcolor{red}{0.755}\cr

\bottomrule
\end{tabular}
 \caption{Quantitative comparison with state-of-the-art methods on synthetic datasets (REDS4, SPMCS, UDM10, YouHQ40) and realworld datasets (VideoLQ). The best and second-best results are marked in \textcolor{red}{red} and \textcolor{blue}{blue}, respectively.}\vspace{-0.5cm}
\label{tab:sota}
\end{center}
\end{table*}

Then we conduct attention-based fusion on these three latent features, the attention weight of the $m$-th head can be calculated by,

\begin{equation}\label{eq:10}
A_m=\rm Softmax\it(\frac{Q_mK_m^{\rm T}}{\sqrt{d_k}}),
\end{equation}
\noindent where
\begin{equation}\label{eq:11}
\begin{aligned}
  & Q_m=W^Q_m[Z^w_{i-1},Z^L_{i},Z^w_{i+1}],  \\
  & K_m=W^K_m[Z^w_{i-1},Z^L_{i},Z^w_{i+1}].  \\
\end{aligned}
\end{equation}
\noindent $d_k$ denotes the dimension of the key vector $K_m$. Then the fused latent feature can be calculated by,
\begin{equation}\label{eq:12}
Z_i^f=\frac{\tilde{Z}^w_{i-1}+\tilde{Z}^L_{i}+\tilde{Z}^w_{i+1}}{3}M_i^f\downarrow_{\times n}+(1-M_i^f\downarrow_{\times n})Z^L_{i},
\end{equation}
\noindent where $\downarrow_{\times n}$ denotes $n$ times downsampling,
\begin{equation}\label{eq:13}
[\tilde{Z}^w_{i-1},\tilde{Z}^L_{i},\tilde{Z}^w_{i+1}]= \frac{\sum_{i=m}^{N_h}A_m}{N_h}[Z^w_{i-1},Z^L_{i},Z^w_{i+1}],
\end{equation}
\noindent and 
\begin{equation}\label{eq:14}
\begin{aligned}
M_i^f=\begin{cases}
 &1 \textbf{ if } e^{-\alpha(\left \|I^w_{i-1}-I^L_{i}\right \|_1+\left \|I^w_{i+1}-I^L_{i}\right \|_1)}> \mu,\\
 &0 \textbf{ if } e^{-\alpha(\left \|I^w_{i-1}-I^L_{i}\right \|_1+\left \|I^w_{i+1}-I^L_{i}\right \|_1)}\le \mu.
\end{cases}
\end{aligned}
\end{equation}

\noindent $N_h$ is the number of the attention heads and $\mu$ is the fusion threshold. $M_i^f$ denotes a hard mask calculated from the warping error between adjacent frames that can restrict the MFF module to do frame fusion only in aligned regions and $\alpha$$>$$0$ is a scaling factor. It is worth noting that we do not map the input latent feature into representation
subspace, as did in vanilla multi-head self-attention \cite{vaswani2017attention}. Such a design allows the fused latent features to still be in the original VAE latent space, thus making better use of pre-training weights.

\subsection{Loss Function}
In order to improve the temporal consistency of the synthetic video, we additionally introduce warp loss defined in \cite{lai2018learning}:
\begin{equation}\label{eq:15}
\mathscr{L}_{warp} = M_i\left \|\hat{I}^H_i-\rm Warp\it(I^H_{i-1},OF^H_{b,i-1})\right \|_2,
\end{equation}
\noindent where
\begin{equation}\label{eq:16}
M_i = e^{-\alpha(\left \|I^H_i-\rm Warp\it(I^H_{i-1},OF^H_{b,i-1}\right \|_1)}.
\end{equation}
\noindent The function of $M_i$ is the same as $M_i^f$, e.g., the loss is only calculated in the region where the warp alignment is accurate.

Given the above defined generator loss $\mathscr{L}_{G}$, focal mean square error loss $\mathscr{L}_{fmse}$, the commonly used perceptual loss $\mathscr{L}_{lpips}$ and the warp loss $\mathscr{L}_{warp}$, the total loss for training OS-DiffVSR is defined as,
\begin{equation}\label{eq:17}
\mathscr{L}_{total} = \omega_1\mathscr{L}_{G}+\omega_2\mathscr{L}_{fmse}+\omega_3\mathscr{L}_{lpips}+\omega_4\mathscr{L}_{warp},
\end{equation}
\noindent where $\omega_1$, $\omega_2$, $\omega_3$, and $\omega_4$ are empirically set to 1, 1, 2, and 2, respectively.

%% file: sec/4_experiments.tex
\section{Experiments}
\subsection{Datasets and Implementaiton Details}

\textbf{Datasets.} We use REDS dataset \cite{nah2019ntire} as our training set, which contains 266 videos with resolution 1280$\times$720. Following OSEDiff \cite{wu2025one}, we employ the degradation process in RealESRGAN \cite{wang2021real} to generate low-quality images for 4$\times$ super-resolution setting. For evaluation, we test on four synthetic datasets (i.e., REDS4 \cite{nah2019ntire}, UDM10 \cite{tao2017detail}, SPMCS \cite{yi2019progressive} and YouHQ40 \cite{zhou2024upscale}) and a real-world dataset (i.e., VideoLQ \cite{chan2022investigating}). According to other mainstream video super-resolution works, during the inference, we generate LR for systhetic datasets with the degradation from RealBasicVSR \cite{chan2022investigating}.

\noindent\textbf{Training Details.} We implement our method using PyTorch and train the model with a batch size of 8 for 150 epochs. During the training, HR images are randomly cropped into 512$\times$512 patches and then utilized to generate corresponding LR images. To train the generator part, we use AdamW \cite{loshchilov2017decoupled} optimizer with initial learning rate of 1e-5 and weight decay of 1e-8. The LoRA rank is set to 32 by default. RAFT \cite{teed2020raft} is used to generate the optical flow to warp the previous and 
successor LR frames. We use SGD optimizer to synchronously train the DINOv2-S \cite{oquab2023dinov2} discriminator. The SGD optimizer will warm up 500 iterations to reach the learning rate of 5e-4. The temperature coefficient $\tau$ and the fusion threshold $\mu$ is set to 100 and 0.4, respectively.

\noindent\textbf{Evaluation Metrics.} Frame quality is measured by both fidelity-based metric (PSNR) and perceptual-driven metrics (CLIP-IQA \cite{wang2023exploring}, MUSIQ \cite{ke2021musiq}, MANIQA \cite{yang2022maniqa}, NIQE \cite{zhang2015feature}). Overall video quality is measured by DOVER \cite{wu2023exploring} and the temporal consistency is evaluated by the warping error($E^{*}_{warp}$) \cite{lai2018learning}.

\begin{figure*}[ht]

  \centering
  \includegraphics[width=\linewidth]{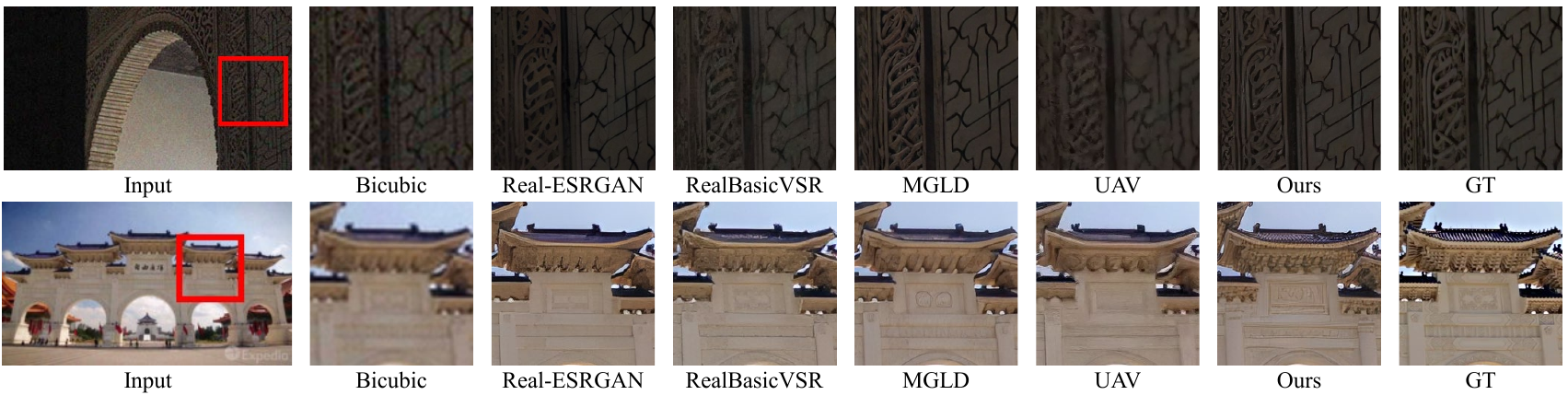}
  \caption{Qualitative comparisons on synthetic LR videos from UDM10 and SPMCS datasets. OS-DiffVSR can recover more realistic and clear architectural details (zoom-in for better view).}\vspace{-0.3cm}
 \label{fig:synthetic}
\end{figure*}
\begin{figure*}[ht]

  \centering
  \includegraphics[width=\linewidth]{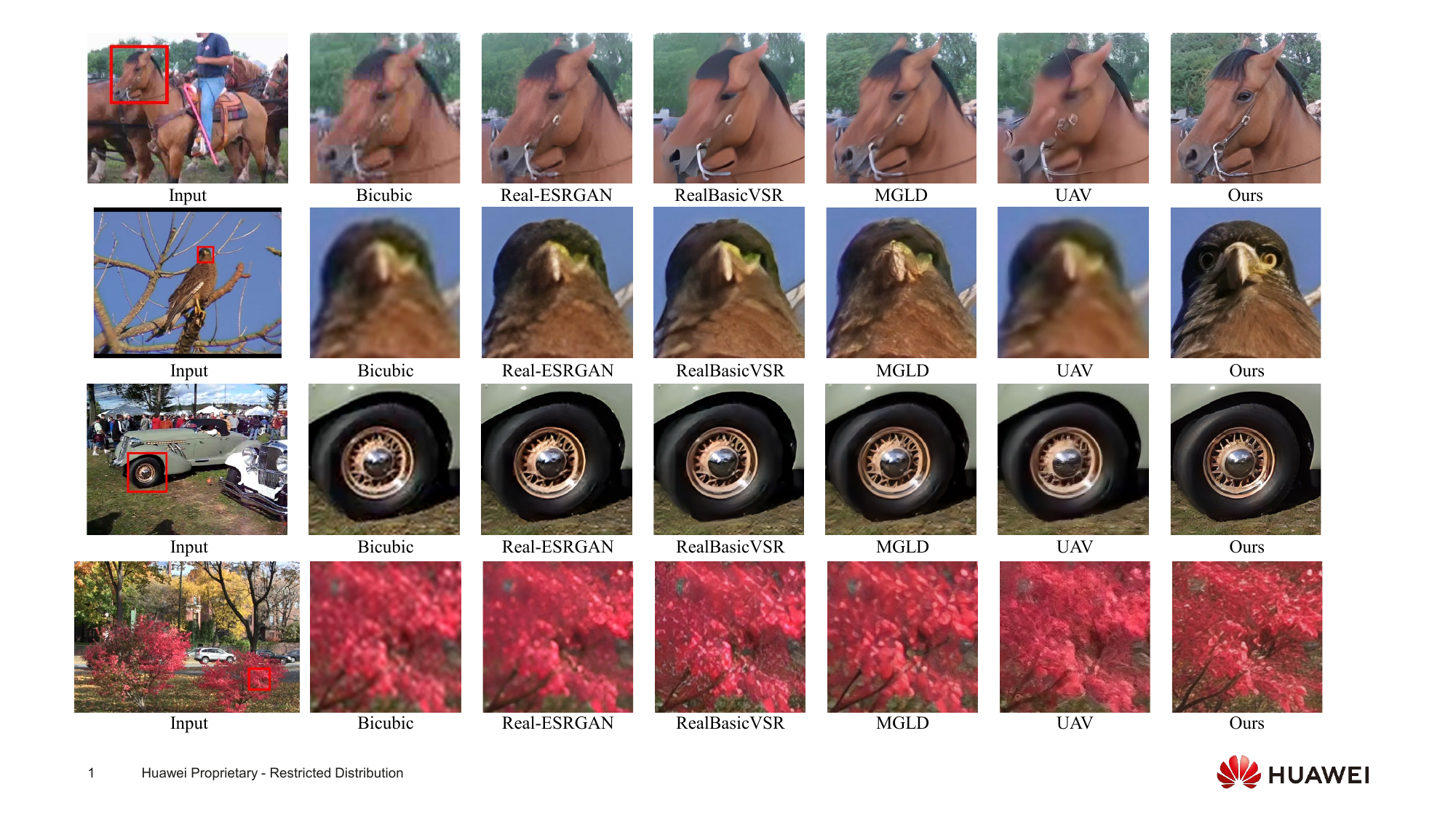}
  \caption{Qualitative comparisons on real-world LR videos from VideoLQ dataset. OS-DiffVSR can can synthesize animals, plants and objects with rich details (zoom-in for better view).}\vspace{-0.4cm}
 \label{fig:real}
\end{figure*}

\begin{table*}[t]
  \centering
\begin{threeparttable}

    \begin{tabular}{c|cccc}
    \toprule
   Settings &PSNR$\uparrow$&CLIP-IQA$\uparrow$&
     MUSIQ$\uparrow$&MANIQA$\uparrow$\cr
    \midrule
    Baseline&\textbf{26.414}&0.530&62.158&0.566\cr
    Vanilla GAN&\textcolor{green}{24.597 }&0.598&\textbf{67.834}&\underline{0.592}\cr
    AFAT (class token)&\underline{25.971}&0.576&63.771&0.578\cr
    AFAT (pacth token)&25.590&\textbf{0.612}&66.115&0.591\cr
    AFAT (pacth token) + FMSE &25.503&\underline{0.606}&\underline{66.529}&\textbf{0.594}\cr
    \bottomrule
    \end{tabular}
    \caption{Evaluation of the adjacent frame adversarial training paradigm. The videos synthesized under the \textcolor{green}{vanilla GAN} training paradigm have serious pseudo-textures (see supplementary materials for examples).
}\vspace{-0.3cm}\label{tab:gan}
    \end{threeparttable}
\end{table*}

\begin{table}[t]\small
  \centering
\begin{threeparttable}

    \begin{tabular}{c|ccc}
    \toprule
   Settings &NIQE$\downarrow$&DOVER$\uparrow$&
     $E^{*}_{warp}\downarrow$\cr
    \midrule
    OSEDiff           &4.049&0.780&1.917\cr
    \midrule
    Baseline          &\underline{3.766}    &0.761&2.003\cr
    MFF&3.868&\underline{0.784}&\underline{1.843}\cr
    MFF+Warp Loss&\textbf{3.727}&\textbf{0.785}&\textbf{1.805}\cr
    \bottomrule
    \end{tabular}
    \caption{Evaluation of the multi-frame adversarial training paradigm.  }\label{tab:fusion}
    \end{threeparttable}
\end{table}

\begin{table}[t]\small
  \centering
\begin{threeparttable}

    \begin{tabular}{c|cccc}
    \toprule
   $\mu$ &CLIP-IQA$\uparrow$&MUSIQ$\uparrow$&
     DOVER$\uparrow$&$E^{*}_{warp}\downarrow$\cr
    \midrule
    1.0&0.607&64.981&0.745&1.862\cr
    \midrule
    0.8&0.599&65.206&0.766&1.830\cr
    0.6&\textbf{0.629}&\underline{66.135}&0.772&\textbf{1.744}\cr
    0.4&0.619&66.072&\textbf{0.785}&1.805\cr
    0.2&\underline{0.626}&\textbf{66.264}&\textbf{0.785}&1.801\cr
    0.0&0.618&66.123&\underline{0.781}&\underline{1.792}\cr
    \bottomrule
    \end{tabular}
    \caption{Evaluation of different fusion threshold $\mu$. $\mu$$=$$1.0$ means the MFF module is not introduced.}\label{tab:fusion thres}\vspace{-0.3cm}
    \end{threeparttable}
\end{table}

\subsection{Comparison with State-of-the-Arts}
We compare OS-DiffVSR with several state-of-the art 
 VSR methods, including Real-ESRGAN \cite{wang2021real}, SD $\times$4 Upscaler \cite{Rombach_2022_CVPR}, RealBasicVSR \cite{chan2022investigating}, MGLD \cite{yang2024motion}, VEnhancer \cite{he2024venhancer}, and Upscale-a-video (UAV) \cite{zhou2024upscale}.
\vspace{-0.3cm}
\paragraph{Quantitative Evaluation.} We evaluate our method on five datasets, including four synthetic degradation datasets and one real-world dataset. As shown in Table~\ref{tab:sota}, CNN-based methods, such as RealBasicVSR and RealESRGAN, tend to over-smooth when dealing with high-frequency degradation (e.g., noise), so they usually achieve a higher PSNR than diffusion-based methods \cite{li2025diffvsr}. Besides, PSNR are primarily designed to measure pixel-level fidelity, while our method focuses on perceptual quality. For the perception-driven metrics, OS-DiffVSR achieves the best performance on almost all datasets with much less inference time than existing multi-step diffusion-based methods (e.g., MGLD, UAV, and VEnhancer). It is worth noting that our method achieves best performance on all metrics on the real-world dataset VideoLQ, which prove that OS-DiffVSR can cope well with various degradations in real-world videos. Besides, We also achieve competitive results on the DOVER, which is the latest video quality metric.
\vspace{-0.3cm}
\paragraph{Qualitative Evaluation.} We present visualization results on synthetic datasets (i.e., UDM10 and SPMCS) in Figure~\ref{fig:synthetic} and a real-world dataset in Figure~\ref{fig:real} to further demonstrate the effectiveness of OS-DiffVSR. The results shows that our method can effectively remove various types of degradation and restore clear and realistic details. As shown in Figure~\ref{fig:synthetic}, OS-DiffVSR can generate more realistic and reasonable architectural details. The results on real-world datasets show that the good generalization ability of OS-DiffVSR enables it to effectively deal with various complex degradations in the real world. More visual comparisons can be found in the supplementary materials.
\subsection{Ablation Study}

We conduct ablation experiments on UDM10 dataset \cite{tao2017detail} to validate the effectiveness of different components of OS-DiffVSR, including the adjacent frame adversarial training paradigm and multi-frame fusion module. More detailed ablation experiments can refer to supplementary materials.

\paragraph{Evaluation of Adjacent Frame Adversarial Training.} We evaluate the effectiveness of the proposed AFAT and the focal MSE loss in Table~\ref{tab:gan}. The baseline setting stands for only utilizing the vanilla MSE loss and the perceptual loss for fine-tuning OSEDiff on our training set. The results show that the perceptual-driven metrics can be greatly improved when the AFAT is introduced. In particular, although the vanilla adversarial training paradigm (i.e., Vanilla GAN) achieves comparable results in perceptual-driven metrics to AFAT, its PSNR significantly dropped by 1.817dB compared to the baseline, which is reflected in the serious pseudo-textures in the synthetic Videos (some relevant examples can be found in the supplementary material). Besides, we also found that utilizing the patch token of DINOv2 to conduct patch-level adversarial training can achieve better performance than using the class token, and the FMSE loss can further boosting the final performance.


\paragraph{Evaluation of Multi-Frame Fusion Module.} We evaluate the effectiveness of MFF module in Table~\ref{tab:fusion}. The baseline setting denotes fine-tuning OSEDiff under AFAT paradigm (i.e., the last row of Table~\ref{tab:gan}). The results show that the proposed MFF module can significantly improve the overall video quality and temporal consistency at the cost of a subtle drop of single frame quality. Besides, introducing the warp loss can further reduce the warp error that measures the temporal consistency.

\vspace{-0.3cm}
\paragraph{Studies on the Fusion Threshold.} We conducted comprehensive studies on the effectiveness of the fusion threshold $\mu$ in Table~\ref{tab:fusion thres}. It is worth noting that $\mu$$=$$0$ is equivalent to not using the MFF module, so it can serve as a baseline for comparison. The results show that as the fusion threshold decreases (i.e., the fusion area expands), the overall video quality and temporal consistency of the method show an overall upward trend. However, a too small $\mu$ will lead to the fusion of more misaligned information. For example, the visual quality degrades when $\mu$$=$$0$. Therefore, we choose $\mu$$=$$0.4$ as our default setting.

Some works have observed that despite the wide use of WE for temporal consistency evaluation, it may not be able to faithfully reflect the real human perception on a video. For example, WE could be easily attacked by blurring the
sequence, resulting much better WE scores but lower video quality \cite{yang2024motion}. In Table~\ref{tab:fusion thres}, we can observe that there is a trade-off between visual quality and temporal consistency. The more details the synthetic video has, the worse the temporal consistency of the video is.

%% file: sec/5_conclusion.tex
\section{Conclusion}
We propose OS-DiffVSR, a one-step diffusion-based video super-resolution method in this work. For improving the perceptual quality, we devise a novel adjacent frame adversarial training paradigm, which can provide more direct guidance for generator by introducing previous frame as a reference. Besides, we introduce a multi-frame fusion module to utilize the information from adjacent frame, thus improving the temporal consistency and the overall video quality. Extensive experiments demonstrate that OS-DiffVSR can achieve state-of-the-art performance with much lower inference time than existing diffusion-based methods. The visualization results show that OS-DiffVSR can generate high-detail and realistic videos.